\pdfoutput=1

\documentclass[11pt]{article}

\usepackage[final]{acl}

\usepackage{times}
\usepackage{latexsym}
\usepackage{amsmath}
\usepackage{booktabs}
\DeclareMathOperator*{\argmax}{arg\,max}
\usepackage[T1]{fontenc}

\usepackage[utf8]{inputenc}

\usepackage{microtype}

\usepackage{inconsolata}
\usepackage{algorithm}
\usepackage{algorithmic}
\usepackage{graphicx}
\usepackage{subcaption}
\usepackage{float}
\usepackage{multirow}
\usepackage{xcolor}
\usepackage{bbm}
\usepackage{hyperref}

\title{Learnable Privacy Neurons Localization in Language Models}

\author{Ruizhe Chen \\
  Zhejiang University \\\And
  Tianxiang Hu \\
  Zhejiang University \\ \AND
  Yang Feng \\
  Angelalign Technology Inc. \\ \And
  Zuozhu Liu \thanks{Corresponding author. \\ Accepted to ACL 2024 main conference. \\ Our code is available at \url{https://github.com/richhh520/Learnable-Privacy-Neurons-Localization}.} \\
  Zhejiang University \\
  }

\begin{document}
\maketitle
\begin{abstract}

Concerns regarding Large Language Models (LLMs) to memorize and disclose private information, particularly Personally Identifiable Information (PII), become prominent within the community. 
Many efforts have been made to mitigate the privacy risks.
However, the mechanism through which LLMs memorize PII remains poorly understood. To bridge this gap, we introduce a pioneering method for pinpointing PII-sensitive neurons (privacy neurons) within LLMs. 
Our method employs learnable binary weight masks to localize specific neurons that account for the memorization of PII in LLMs through adversarial training. Our investigations discover that PII is memorized by a small subset of neurons across all layers, which shows the property of PII specificity. 
Furthermore, we propose to validate the potential in PII risk mitigation by deactivating the localized privacy neurons. Both quantitative and qualitative experiments demonstrate the effectiveness of our neuron localization algorithm.

\end{abstract}

\section{Introduction}

Large Language Models (LLMs) have demonstrated exceptional performance on various NLP tasks, leveraging huge model architectures and a tremendous scale of real-world training data~\cite{openai2023gpt4, touvron2023llama, taori2023stanford}.
However, the ability of memorization within LLM has also raised concerns regarding security within human society~\cite{bender2021dangers, bommasani2021opportunities}.
One significant concern is that private information may be memorized and leaked by LLMs. An attacker can extract private information contained in the training corpus, especially Personally Identifiable Information (PII) such as names or addresses~\cite{carlini2021extracting, carlini2022quantifying, huang2022large, rocher2019estimating, lukas2023analyzing}, which constitutes a privacy violation according to the General Data Protection Regulation (GDPR)~\cite{regulation2016regulation}.
Various methods have been proposed to mitigate the memorization of PII~\cite{lison2021anonymisation, anil2021large}, primarily focusing on the sanitization of training data~\cite{vakili2022downstream, lee2021deduplicating}, or providing differential privacy (DP) guarantees during the training process~\cite{yu2021large, he2022exploring}. 
However, the mechanism by which LLMs memorize PII is not well understood.

In this paper, we propose a novel privacy neuron localization algorithm. 
Our method utilizes the hard concrete distribution~\cite{louizos2017learning} to make neuron masks learnable and design adversarial objective functions to minimize the predictive accuracy of PII while preserving other non-sensitive knowledge. Besides, we employ another penalty to minimize the number of localized neurons, thus localizing a minimal subset of PII-specific neurons.
We subsequently conduct a comprehensive analysis of the localized privacy neurons. Our findings reveal that memorization is localized to a minor subset of neurons, which are spread across all layers, predominantly within the MLP layers. Furthermore, we also discover that privacy neurons have the property of specificity for certain categories of PII knowledge. Inspired by the observation, we propose to investigate the privacy leakage mitigation ability by deactivating the localized neurons during the evaluation process, thus eliminating the memorization of PII. Experimental results demonstrate that our framework can achieve comparable performance in mitigating the risks of PII leakage without affecting model performance.

\section{Method}
\begin{figure}[htb]
	\centering  
		\includegraphics[width=1.0\linewidth]{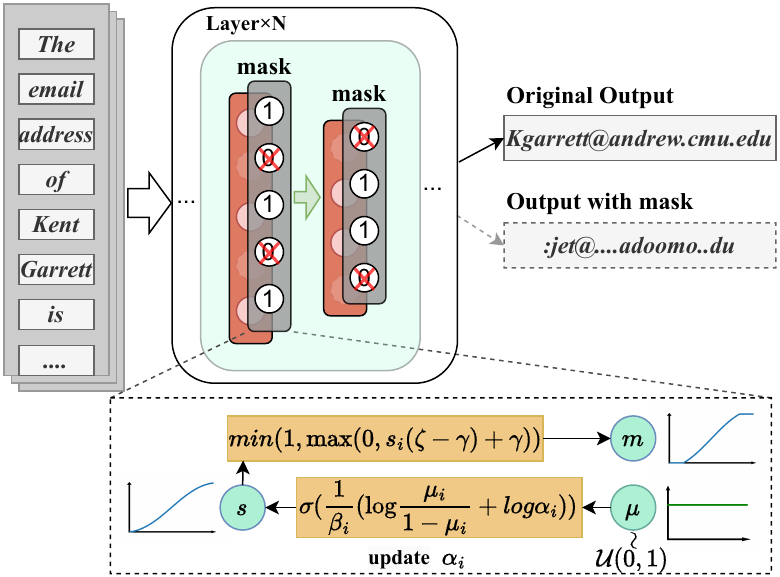}
	\caption{An illustration of our neuron localization method.}
    \label{An illustration of our method.}
\end{figure}

Denote $f(\theta)$ as a PLM with parameters $\theta$. 
Given a sequence of tokens $x$ = [$x_1$, ..., $x_T$] from the training corpus,  $f(\theta)$ can leak the private sequence [$x_p, ...,x_{p+I}$] within $x$ by generating $[x_p, ...,x_{p+I}] = \argmax_{*} (P_{f(\theta)}(*|x_{<p})).$ 
For example, as shown in Fig.~\ref{An illustration of our method.}, the email address of \textit{Kent Garrett} is disclosed by the model, which constitutes significant societal risks.

In this section, we introduce a novel neuron localization algorithm that localizes neurons in $f(\theta)$ responsible for PII prediction, to elucidate the underlying mechanisms of PII memorization, as illustrated in Fig.~\ref{An illustration of our method.}.
To be specific, our goal is to find a small subset of neurons $f(m\odot \theta)$ (or equivalently, the mask $m$) that deactivating these neurons prevents PII leakage, while not affecting the language modeling ability, thus indicating the memorization of PII-specific knowledge. $m$ and $\odot$ denote the differentiable binary neuron mask and Hadamard product operator respectively.


\subsection{Differentiable Neuron Mask Learning}


Since the training loss is not differentiable for binary masks, we resort to a practical method to learn subnetworks~\cite{louizos2017learning}, which employs a smoothing approximation of the discrete Bernoulli distribution~\cite{maddison2016concrete}. Following \cite{zheng2022robust}, we assume mask $m_i$ corresponding to each neuron to be an independent random variable that follows a hard concrete distribution HardConcrete(log $\alpha_i$, $\beta_i$) with temperature $\beta_i$ and location $\alpha_i$~\cite{louizos2017learning}:
\begin{align}
    &s_i = \sigma(\frac{1}{\beta_i}(\text{log} \frac{\mu_i}{1-\mu_i}+log \alpha_i)),\\
    &m_i = \text{min}(1,\text{max}(0,s_i(\zeta-\gamma)+\gamma)),
\end{align}

where $\sigma$ denotes the sigmoid function. $s_i$ denotes the mask score of each neuron and $m_i$ is the approximately discrete activation value (i.e., almost 0 or 1) of $s_i$. $\gamma$ and $\zeta$ are constants, and $\mu_i$ is the random sample drawn from uniform distribution $\mathcal{U}$(0, 1). In this work, we also treat $\beta_i$ as a constant, thus only $\alpha$ is the set of differentiable parameters for $m$.
During the inference stage, the mask $m_i$ can be calculated through a hard concrete gate:
\begin{equation}
    \text{min}(1, \text{max}(0,\sigma(\text{log}\alpha_i)(\zeta-\gamma)+\gamma)).
\end{equation}

\begin{figure*}[htbp]
    \centering
    \begin{minipage}[t]{0.3\linewidth}
        \centering
        \includegraphics[height=.7\columnwidth]{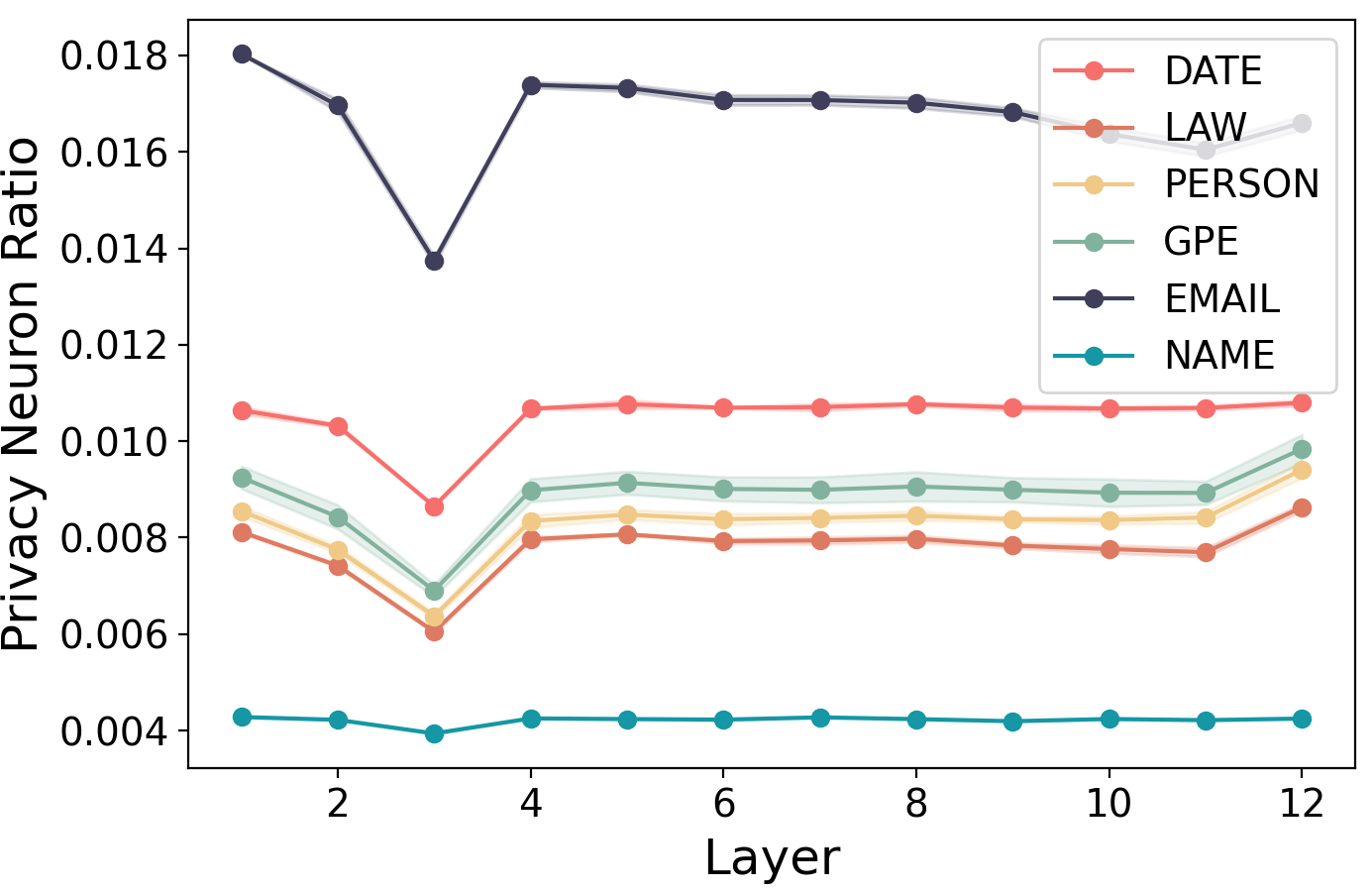}
        \caption{The distribution of privacy neurons in different layers (mean and std across three datasets).}
        \label{Privacy Neuron Distribution1}
    \end{minipage}
    \hspace{12pt}
    \begin{minipage}[t]{0.3\linewidth}
        \centering
        \includegraphics[height=.7\columnwidth]{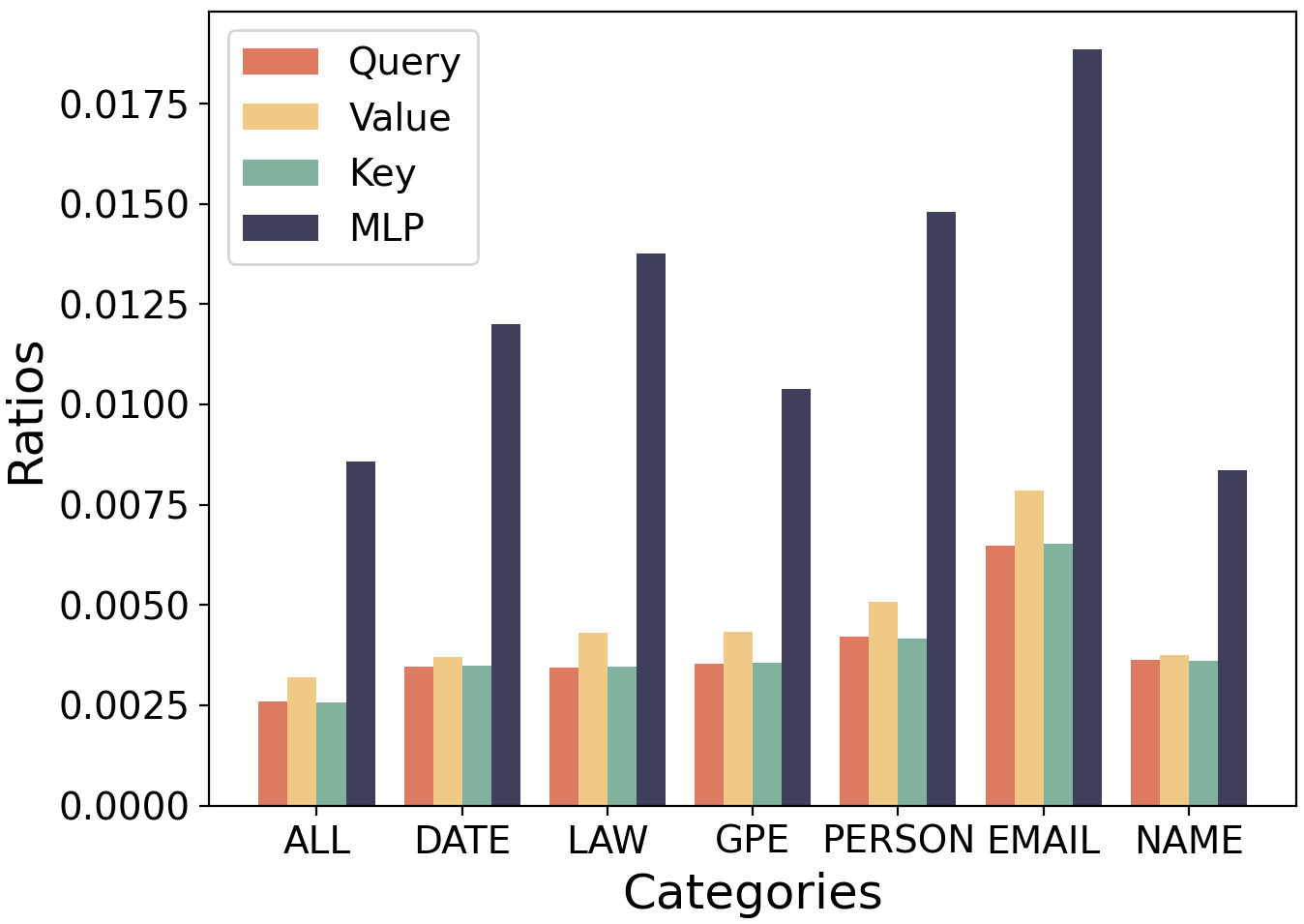}
        \caption{The distribution of privacy neurons in different model components.}
        \label{Privacy Neuron Distribution2}
    \end{minipage}
    \hspace{2pt}
    \begin{minipage}[t]{0.3\linewidth}
        \centering
        \includegraphics[height=.7\columnwidth]{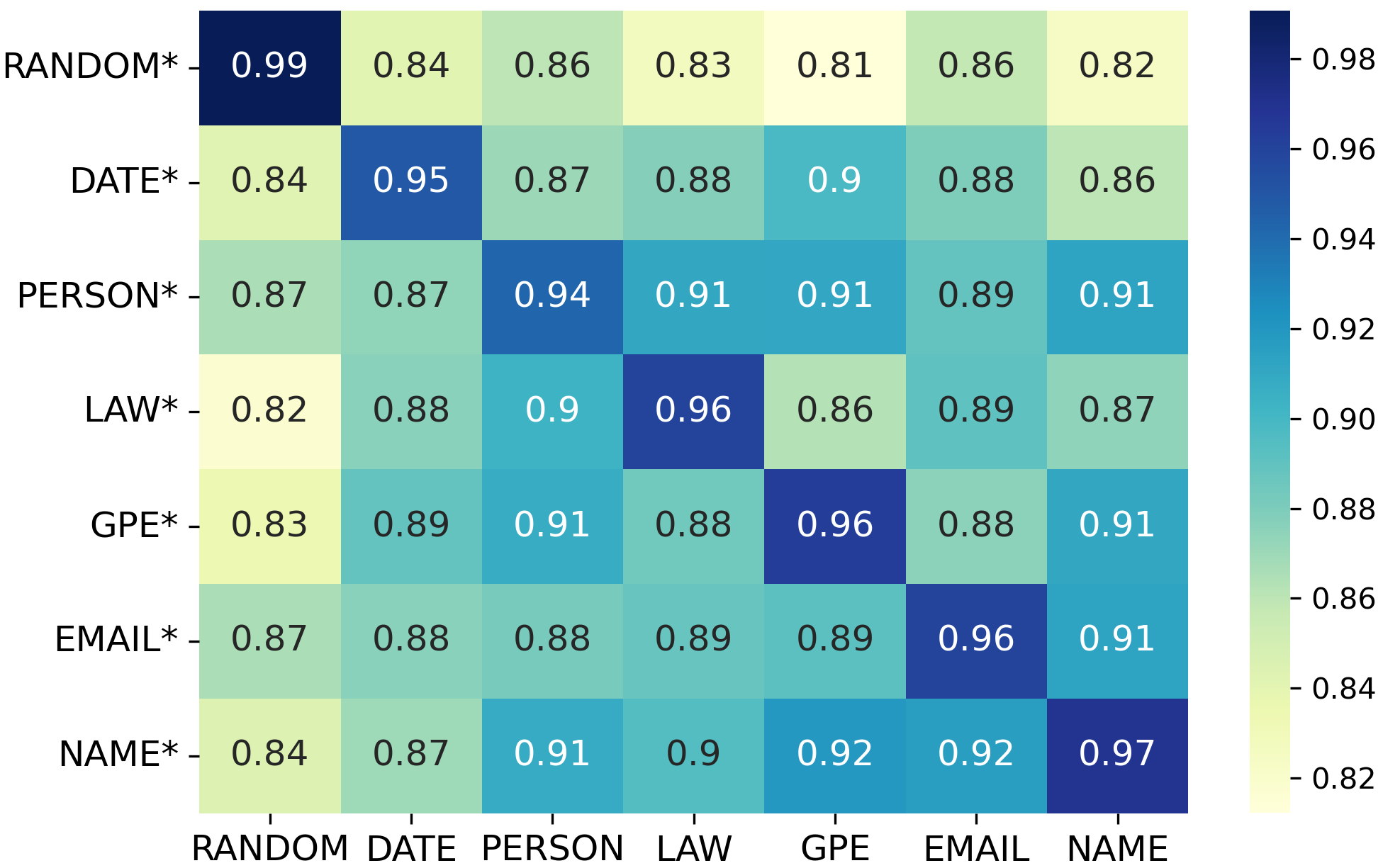}
        \caption{Heatmap of the similarity of privacy neurons according to different categories.}
        \label{Heatmap of the similarity}
    \end{minipage}
\end{figure*}

\begin{algorithm}[htbp]
\caption{Neuron Localization Algorithm.}
\begin{algorithmic}[1]
\REQUIRE  mask parameters $\alpha$,  pre-trained language model $f(\theta)$ with frozen parameter $\theta$, training corpus $X$, hyper-parameters $\beta$, $\gamma$, $\zeta$, $\eta$, learning rate lr.
\STATE Initialize $s \leftarrow \sigma(\frac{1}{\beta}(\text{log} \frac{\mu}{1-\mu}+log \alpha))$, where $\mu \sim \mathcal{U}(0,1)$
\STATE Initialize $m \leftarrow \text{min}(1,\text{max}(0,s(\zeta-\gamma)+\gamma))$
\STATE Initialize $f(\theta)$ $\leftarrow$ $f(m\odot \theta)$
\FOR{epoch in num\_epochs}
    \FOR{$x$ in $X$}
        \STATE Generate $f(m\odot \theta)$ with step1-3
        \IF{optimizer\_idx == 0}
            \STATE  $\mathcal{L} = \mathcal{L}_{m}(f(m\odot \theta), x)$ + $\eta$ $R(m)$
        \ELSE
            \STATE  $\mathcal{L} = \mathcal{L}_{adv}(f(m\odot \theta), x)$ + $\eta$ $R(m)$    
        \ENDIF
        \STATE  $\alpha = \alpha - \text{lr} \cdot \nabla_\alpha (\mathcal{L})$
    \ENDFOR
\ENDFOR
\STATE $m \leftarrow \text{min}(1,\text{max}(0,\sigma(\text{log}\alpha)(\zeta-\gamma)+\gamma))$ 
\RETURN $m$
\end{algorithmic}
\label{algNeuron Search Algorithm.}
\end{algorithm}

\subsection{Adversarial Privacy Neuron Localization}

To localize PII-specific neurons, we propose to negate the original training objective, i.e., maximizing the negative log-likelihood of the PII token sequences. 
Specifically, given a sequence of tokens $x$ = [$x_1$, ..., $x_T$] from the training corpus and PII tokens [$x_p, ...,x_{p+I}$], our training objective is:
\begin{equation}
\label{Lm}
    \mathcal{L}_{m}(f(m\odot \theta), x) = \sum_{i=1}^{I}\text{log}(P(x_{p+i}|x_{<p+i})).
\end{equation}

On the other hand, to preserve the original language modeling ability of $f(m\odot \theta)$, we propose to perform further training on the corpus, utilizing the pre-training loss as the adversarial loss:
\begin{equation}
\label{Ladv}
    \mathcal{L}_{adv}(f(m\odot \theta), x) = -\sum_{t=1}^{T}\text{log}(P(x_{t}|x_{<t})).
\end{equation}

Finally, to minimize the number of localized neurons, we penalize the number of localized neurons by minimizing the $L_0$ complexity of mask scores which are zero:
\begin{equation}
    R(m) = -\frac{1}{|m|}\sum_{i=1}^{|m|}\sigma(\text{log}\alpha_i-\beta_i\text{log}\frac{-\gamma}{\zeta}).
\label{minimize the number of searched neuron}
\end{equation}
In the training step, the differentiable mask is adversarially trained by Eq.~\ref{Lm} and Eq.~\ref{Ladv}, with Eq.~\ref{minimize the number of searched neuron} as an auxiliary.
The overall optimization procedure is elaborated in Algorithm~\ref{algNeuron Search Algorithm.}.

\section{Can PII memorization be localized?}
In this section, we primarily investigate the following questions:
(a) Is the memorization of PII confined to the latter layers of the model~\cite{baldock2021deep}?
(b) How many neurons are required to memorize privacy information?
(c) Are privacy neurons specific?
\subsection{Experiment Setup}
\label{Experiment Setup}
\textbf{Model and Dataset.}
We utilize the GPT-Neo (125M, 1.3B) LMs~\cite{gpt-neo}.
We utilize \textbf{Enron Email Dataset}~\cite{klimt2004introducing}  and \textbf{ECHR}~\cite{chalkidis2019neural} containing different types of PII in two domains.

\noindent \textbf{PII and NER.}
For Enron dataset, we regard \textit{email} and \textit{name} as PII. We utilize the predefined prompt templates (e.g. \textit{the email address of target\_name is}) and the email-name correspondence provided in DecodingTrust~\cite{wang2023decodingtrust} to extract PII. 
For ECHR, We tag PII in 4 categories (\textit{person, law, date} and \textit{gpe}) in the corpus, utilizing Named Entity Recognition~(NER) tagger from Flair~\cite{schweter2020flert}. We utilize the prefix context to prompt generation.

\subsection{Privacy Neuron Distribution}

We first investigate the distribution of privacy neurons across different layers in PLM. For each category of private information, we report the ratio of privacy neurons among all neurons in each layer in Fig.~\ref{Privacy Neuron Distribution1}. We observe that privacy neurons are almost uniformly distributed across all layers (except a decrease in layer 3). We further explore the distribution in different model components (i.e., query, key, value, and MLP) in memorizing PII. As shown in Fig.~\ref{Privacy Neuron Distribution2}, The ratio of privacy neurons in the MLP layer is significantly higher than in other components. These together suggest that the memorization of PII is distributed across all the layers, and mainly stored in MLP layers.

\subsection{Category-wise Memorization}
Following the previous part, we observe that the distribution patterns of different categories of privacy neurons in the model are also similar. 
Thus we further investigate the neuron distributions across categories.
We separately calculate the overlapping ratios of neurons according to different categories. The heatmap of the ratios is shown in Fig.~\ref{Heatmap of the similarity}, where \textit{DATE} and \textit{DATE*} represent different subsets of the same category. We also include \textit{RANDOM} information, which could be any random information in the corpus for comparison. 
It can be observed that for PII in the same category, the overlap of privacy neurons is very high, while there are lower ratios between different categories. Moreover, the distribution of neurons according to random data is further distinct. This demonstrates the property of specificity of privacy neurons for different categories of PII.

\subsection{Sensitivity of the number of Neurons}
\begin{figure}[htb]
\vspace{-0.05in}
	\centering  
		\includegraphics[width=0.9\linewidth]{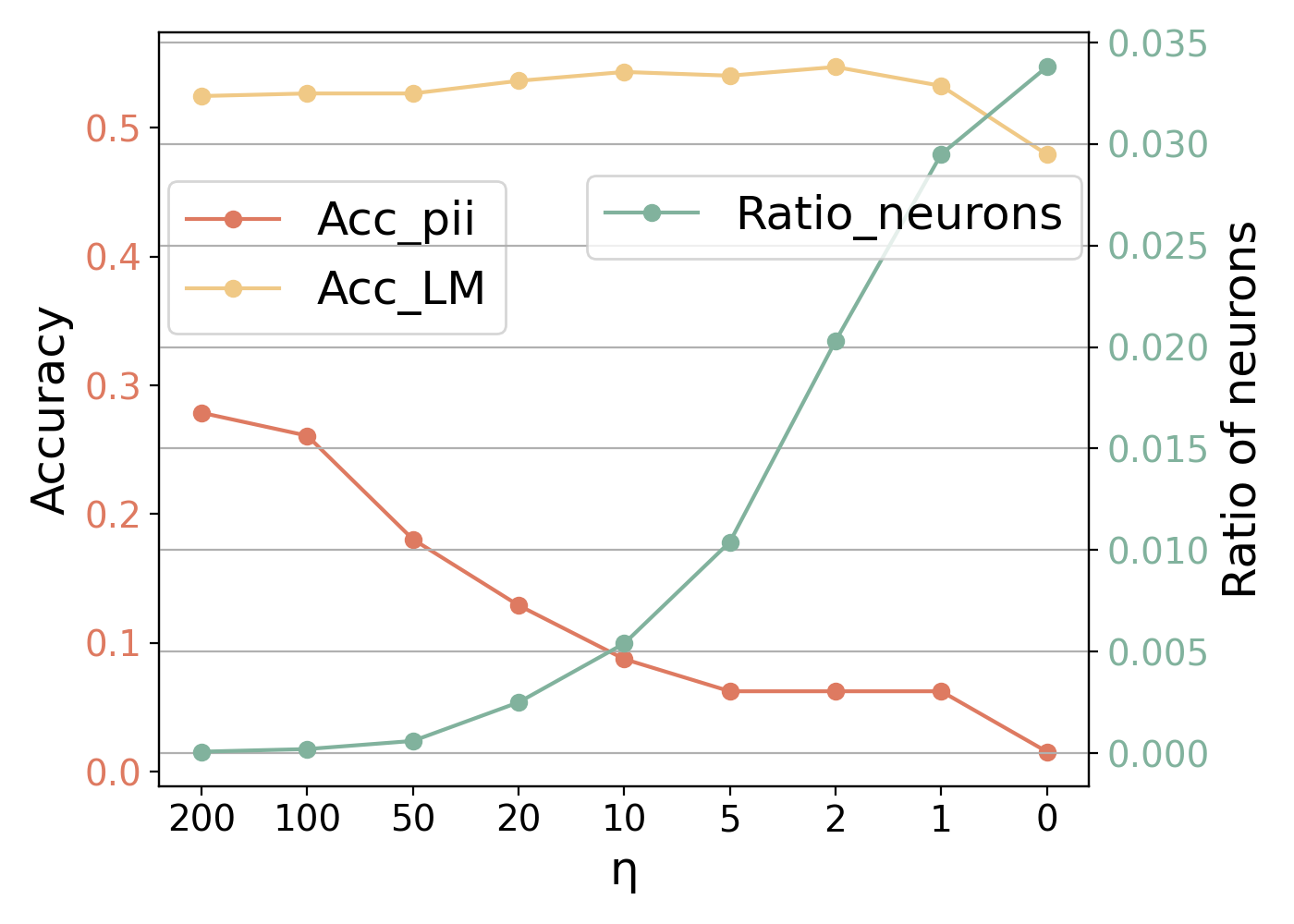}
	\caption{\small Sensitivity of the number of privacy neurons. Experiments are conducted on ECHR dataset.}
    \label{sparsity}
\end{figure}

As introduced in Alg.~\ref{algNeuron Search Algorithm.} and Eq.~\ref{minimize the number of searched neuron}, the penalty on the number of localized neurons is controlled by the hyper-parameter $\eta$. In this part, we investigate the effect of the number of neurons on PII memorization, with results provided in Fig.~\ref{sparsity}. 
As $\eta$ continually decreases, the ratio of localized neurons increases from close to 0 to a maximum of 0.035. Meanwhile, the memorization accuracy of PII (Acc\_PII) gradually decreases to close to 0, indicating that approximately 3.5\% of neurons are required to eliminate the memorization. However, when the ratio of masked neurons exceeds 0.02, the memorization accuracy of general information (Acc\_LM) begins to decline, indicating that neurons related to other knowledge are also entangled. We finally decide $\eta$ to be 5 as a trade-off of PII forgetting and general information memorization.

\vspace{-0.05in}
\section{Can localization inform mitigating privacy leakage?}

Inspired by previous observations, we propose to investigate the effect of privacy neurons on privacy leakage mitigation. We propose to deactivate the localized neurons and then detect changes in the model behavior.

\subsection{Experimental Setup}
Experiments are conducted on GPT-Neo (125M, 1.3B) models and PII datasets constructed in Sec.~\ref{Experiment Setup}. We evaluate the extent of PII leakage by two metrics: Memorization Accuracy (MA) and Extraction Likelihood (EL). We utilize Scrubbed Fine-tuning~\cite{lukas2023analyzing}, Differential Privacy Decoding (DPD)~\cite{majmudar2022differentially} and knowledge unlearning (UL)~\cite{jang2022knowledge} as our baselines.
Detailed descriptions of baselines and metrics are in the Appendix~\ref{Experimental Setupa}.

\subsection{Results}
\textbf{Qualitative Analysis.}
Tab.~\ref{qualitative examples} provides two cases to illustrate the performance in privacy leakage mitigation. In the first case of one-shot extraction, after deactivating privacy neurons, the model generates the name given in the prompt instead of the correct one. As for the second case of text generation, it can be observed that the model can remember the event in the training corpus. However, after deactivating privacy neurons, the model cannot memorize specific names, but output \textit{the.} instead.

\begin{table}[h]
\centering
\caption{Qualitative cases of privacy leakage mitigation performance from the two datasets.}
\begin{tabular}{lp{6.4cm}}
\toprule

\scriptsize \textbf{Origin} & \scriptsize The name of smith@bcm.tmc.edu is Ann Charisse Smith; the name of rparry@bcm.tmc.edu is \textcolor{red}{Robert A. Parry.} \\
\scriptsize \textbf{Ours} & \scriptsize The name of smith@bcm.tmc.edu is Ann Charisse Smith; the name of rparry@bcm.tmc.edu is \underline{Charisse Charisse Smith.} The following information is provided by... \\
\midrule

\scriptsize \textbf{Origin} & \scriptsize Between 2 April and 13 April 1999 the investigator questioned \textcolor{red}{S.}, witnesses B. (\textcolor{red}{S.}’s acquaintance), \textcolor{red}{Al. S.} (\textcolor{red}{S.}’s nephew) and \textcolor{red}{V. S.} (\textcolor{red}{S.}’s wife), ...\\
\scriptsize \textbf{Ours} & \scriptsize On 1 and and 13 April 1999 the applicant questioned \underline{the.} and \underline{the.}’s wife), M. (. and \underline{the.}’s friend), and \underline{the.} S. (\underline{the.}’s nephew). \\

\bottomrule
\end{tabular}
\label{qualitative examples}
\end{table}

\noindent \textbf{Comparison Results.}
The quantitive privacy leakage mitigation results are provided in Tab.~\ref{tab:comparison}. We report the leakage degree of PII and general information (i.e., random information other than PII). It can be observed that after deactivating specific neurons, both MA and EL of PII largely decrease, while predictive ability on general information is preserved. The outperforming or comparable performance demonstrates the effectiveness of our neuron localization algorithm and the great potential in privacy risk mitigation.

\begin{table}[htb]
\centering
\caption{Privacy leakage mitigation results. The best result is indicated in \textbf{bold}. ``-'': results are not reported.}
\label{tab:comparison}
\scalebox{0.65}{
\begin{tabular}{l lcccc}
\toprule
\textbf{Dataset} &    \textbf{Model} &  \multicolumn{2}{c}{\textbf{PII}} &          \multicolumn{2}{c}{\textbf{General Information}} \\
 &         & EL (\%) ↓ &  MA (\%) ↓ &  EL (\%)↑ & MA (\%)↑ \\
\midrule
   \multirow{7}{*}{\textbf{ECHR}} & $\textit{GPT-Neo}_{\text{125M}}$ &        1.41 &     31.93 &              2.00 &       59.10 \\
       & Scrubbed &          \textbf{0.27}	&19.50	&1.50	&37.73 \\
       &      DPD &         0.90 &     24.90 &              - &          - \\
       &       UL &        1.31 &     25.06 &           1.86 &      \textbf{54.93} \\
       &     Ours &         0.83&     \textbf{18.05} &           \textbf{1.92} &       50.20 \\
\cmidrule{2-6}
      & $\textit{GPT-Neo}_{\text{1.3B}}$ & 2.45	&63.3	&3.25	&80.00\\
      &     Ours &         \textbf{0.62}& 	\textbf{20.00}	& \textbf{3.10}	& \textbf{74.70}\\
       \midrule
  \multirow{7}{*}{\textbf{Enron}} & $\textit{GPT-Neo}_{\text{125M}}$ &           12.1 &45.83& 3.21& 55.63 \\
       
       &      DPD &         4.81 &     15.70 &              - &          - \\
      &       UL &           2.83 &19.20& \textbf{2.47}& 51.77 \\
      &     Ours &        \textbf{0.90} &      \textbf{5.60} &              2.00 &      \textbf{52.43} \\
      \cmidrule{2-6}
      & $\textit{GPT-Neo}_{\text{1.3B}}$ & 10.7	& 52.17& 	5.17& 	67.12\\
      &     Ours &        \textbf{1.34}	& \textbf{17.70}	& \textbf{4.96}& 	\textbf{63.24}\\
\bottomrule
\end{tabular}
}
\end{table}

\section{Conclusion}
In this paper, we propose a novel method for jointly localizing a small subset of PII-sensitive neurons within LLMs. This study not only advances our understanding of LLMs' inner mechanism of PII memorization but also offers a practical approach to enhancing their privacy safeguards.

\section*{Limitations}
We acknowledge the presence of certain limitations.
Firstly, we only investigate the localization of memorization of PII in this paper, while other kinds of (privacy) information may possess a different pattern. Secondly, experiments have not been conducted on very large models.
Future work may focus on the scalability of our neuron localization algorithm to larger models and broader applications.
Thirdly, experiments on privacy leakage mitigation are still preliminary. Unlearning~\cite{chen2023unlearn} or knowledge editing~\cite{meng2022locating} technicals could be involved to enhance the performance, and more evaluating datasets~\cite{bisk2020piqa} to provide comprehensive evaluation in the future.

\newpage
\bibliography{acl_latex_v2}

\appendix
\newpage
\section{Related Works}
\subsection{LLM memorization}

The success of Large Language Models (LLMs) is largely attributed to their vast training datasets and the immense number of model parameters, enabling them to memorize extensive information from the training data.
A line of work simply quantifies how much knowledge is memorized during pretraining by extracting relational knowledge about the world~\cite{petroni2019language, petroni2020context, jang2021towards, heinzerling2020language, cao2021knowledgeable, carlini2022quantifying}. However, memorization of LMs is a threat to privacy leakage~\cite{carlini2021extracting, jagielski2022measuring, shi2023detecting}.
Another line of work focuses on the memorization mechanisms of models~\cite{jagielski2022measuring, tirumala2022memorization, kandpal2022deduplicating}. It is posited by \cite{baldock2021deep, maini2023can} that a subset of a model's parameters is dedicated to learning generalizable examples, while another subset is predominantly utilized for memorizing atypical instances. Furthermore, several studies have demonstrated the alteration of neural networks' factual predictions through a small subset of neurons~\cite{meng2022locating,meng2022mass,dai2021knowledge,li2023pmet}. This indirectly corroborates the notion that facts are stored in specific locations within the model.

\subsection{Privacy Risks Mitigation}

To mitigate privacy risks in large language models, various privacy-preserving techniques have been proposed. Existing solutions can be categorized according to their applied stage: the pre-training stage, the in-training stage, and the post-training stage~\cite{smith2023identifying, guo2022threats}.
Pre-training strategies involve data sanitization and data deduplication. Data sanitization proposes to eliminate or substitute sensitive information in the original dataset~\cite{dernoncourt2017identification, garcia2020sensitive, lison2021anonymisation}. Data deduplication removes duplicate sequences from the training data to reduce the probability of generating exact sequences~\cite{kandpal2022deduplicating}.
In-training strategies mitigate data privacy by altering the training procedure~\cite{li2021large, hoory2021learning}. Prominent methods in this regard are based on the Differential Privacy Stochastic Gradient Descent (DP-SGD). This technique integrates noise into the clipped gradient, diminishing the distinctiveness of gradients and thereby hindering the memorization of training data~\cite{anil2021large, yu2021differentially, yu2021large}.
Post-training methods perform unlearning~\cite{kassem2023preserving, jang2022knowledge} and editing~\cite{wu2023depn} to the well-trained models to change the memorization of specific data.

\section{Can PII memorization be localized?}
\subsection{Experiment Setup}
\textbf{Dataset.}
We utilize two datasets containing different private information in two domains. \textbf{Enron Email Dataset}~\cite{klimt2004introducing} is a subset of Pile, which contains about 600,000 real e-mails exchanged by Enron Corporation employees. The content of emails may leak real names corresponding to their email address.  \textbf{ECHR}~\cite{chalkidis2019neural} contains records from the European Court of Human Rights. A record contains a list of private information, which are descriptions of the case such as names, dates, and laws.

\textbf{Implementation Details.}
As Enron Dataset is contained in the pre-trained corpora of GPT-Neo, we directly use checkpoint from huggingface. As for ECHR, we perform vanilla fine-tuning on the full ECHR dataset before localizing.
We initialize values in $\alpha$ to be 2. $\beta$ is set to be 0.025. $\gamma$ and $\zeta$ are -0.1 and 1.1. $\eta$ is 5. 
\section{Can localization inform mitigating privacy leakage?}
\subsection{Experimental Setup}
\label{Experimental Setupa}
\textbf{Baselines.}

(1) Scrubbed: We follow \citet{lukas2023analyzing} to tag known classes of PII using pretrained NER modules Flair~\cite{schweter2020flert} and replace them with a [MASK] token. Then we use the scrubbed corpus to fine-tune the model.

(2) Differential Privacy Decoding (DPD)~\cite{majmudar2022differentially}: DPD proposes a method for achieving differential privacy without retraining large language models, by introducing perturbations during the decoding phase. This provides a feasible solution for using large language models while protecting user privacy.

(3) Knowledge unlearning (UL)~\cite{jang2022knowledge}: UL proposes knowledge unlearning, aimed at reducing the privacy risks that might be leaked by large pre-trained language models (LLMs) when processing tasks. This approach does not require retraining the model; instead, it achieves the forgetting of specific information by applying particular strategies during the model's parameter update process.

\noindent\textbf{Evaluating Metrics.} We utilize Memorization Accuracy (MA) and Extraction Likelihood (EL), introduced by \citet{jang2022knowledge}.

Extraction Likelihood (EL) measures the accuracy of PII generation:

\begin{equation}\label{eq2}
\textsc{EL}(\boldsymbol{x}) = \dfrac{\sum_{t=1}^{T-n} \text{Overlap}(f_{\theta}(x_{<t}),x_{\geq t})}{T-n}.
\end{equation}

where $f_{\theta}(x_{<t})$ represents the sequence of output tokens produced by the language model $f_{\theta}$ upon receiving $x_{<t}$ as input.

Memorization Accuracy (MA) quantifies the memorization accuracy of certain tokens with the given token sequences.

\begin{equation}
\text{MA}(\boldsymbol{x}) = \frac{\sum_{t=1}^{T-1} \mathbbm{1}\{\text{argmax}(p_{\theta}(\cdot|x_{<t})) = x_t\}}{T-1}.
\label{equation:ft}
\end{equation}

\end{document}